\documentclass[10pt,twocolumn,letterpaper]{article}

\usepackage{wacv}
\usepackage{times}
\usepackage{epsfig}
\usepackage{graphicx}
\usepackage{amsmath}
\usepackage{amssymb}
\usepackage{booktabs}
\usepackage{kotex}
\usepackage{multirow}
\usepackage{threeparttable}

\usepackage{pifont}% http://ctan.org/pkg/pifont
\usepackage{xcolor}
\usepackage{pdfrender}
\newcommand{\xmark}{\ding{55}}%
\usepackage{colortbl}
\usepackage{bbm}
\usepackage[accsupp]{axessibility}
\usepackage{float}
\usepackage{makecell}

\newcommand*{\boldcheckmark}{%
  \textpdfrender{
    TextRenderingMode=FillStroke,
    LineWidth=.5pt, % half of the line width is outside the normal glyph
  }{\checkmark}%
}
\newcommand*{\boldxmark}{%
  \textpdfrender{
    TextRenderingMode=FillStroke,
    LineWidth=.5pt, % half of the line width is outside the normal glyph
  }{\xmark}%
}
\newcommand*{\affmark}[1][*]{\textsuperscript{#1}}
\newcommand\blfootnote[1]{%
  \begingroup
  \renewcommand\thefootnote{}\footnote{#1}%
  \addtocounter{footnote}{-1}%
  \endgroup
}

\def\wacvPaperID{1256} % Enter the WACV Paper ID here

\wacvalgorithmstrack   % Uncomment this line if you are submitting to the Algorithms Track.
\wacvfinalcopy % *** Uncomment this line for the final submission

\ifwacvfinal
\usepackage[breaklinks=true,bookmarks=false,hyperfootnotes=false]{hyperref}
\else
\usepackage[pagebackref=true,breaklinks=true,colorlinks,bookmarks=false]{hyperref}
\fi

\begin{document}

%%%%%%%%% TITLE
\title{Dense but Efficient VideoQA for Intricate Compositional Reasoning}

\author{
Jihyeon Lee\affmark[*]\\
Kakao Brain\\
{\tt\small gina.ai@kakaobrain.com}
\and
Wooyoung Kang\affmark[*]\\
Kakao Brain\\
{\tt\small edwin.kang@kakaobrain.com}
\and
Eun-Sol Kim\\
Department of Computer Science, \\Hanyang University\\
{\tt\small eunsolkim@hanyang.ac.kr}
}

\maketitle
\thispagestyle{empty}

%%%%%%%%% ABSTRACT
\begin{abstract}
It is well known that most of the conventional video question answering (VideoQA) datasets consist of easy questions requiring simple reasoning processes. However, long videos inevitably contain complex and compositional semantic structures along with the spatio-temporal axis, which requires a model to understand the compositional structures inherent in the videos. In this paper, we suggest a new compositional VideoQA method based on transformer architecture with a deformable attention mechanism to address the complex VideoQA tasks. The deformable attentions are introduced to sample a subset of informative visual features from the dense visual feature map to cover a temporally long range of frames efficiently. Furthermore, the dependency structure within the complex question sentences is also combined with the language embeddings to readily understand the relations among question words. Extensive experiments and ablation studies show that the suggested dense but efficient model outperforms other baselines.
\end{abstract}

%%%%%%%%% BODY TEXT
\vspace{-0.5cm}
\section{Introduction}
\begin{figure}
    \begin{center}
        \includegraphics[width=\linewidth]{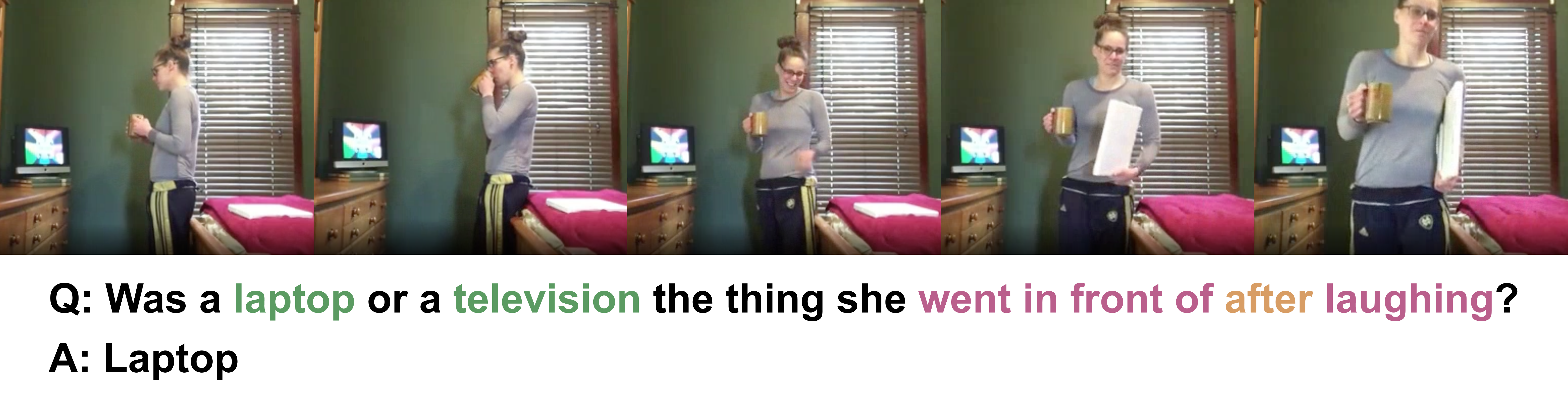}
        \vspace{-0.5cm}
    \end{center}
    \caption{Example of an intricate VideoQA problem. The semantic elements of the video, which consists of characters, their actions, and the relationship between the characters, is continually changing along with temporal axis. Therefore, it is hard to answer the questions requiring understanding the complex semantic structure.}
    \label{fig:vqa_example}
    \vspace{-0.5cm}
\end{figure}

Along with the immense success of deep learning methods to understand the contents of images and text, various applications requiring complex reasoning have been proposed.
Especially, visual question answering (VQA)~\cite{Antol_2015_ICCV} is one of the most important tasks, which asks a diverse set of questions about the visual contents and requires understanding the semantic structures inherent in the contents.
By virtue of the emergence of transformer architectures and their pre-training schemes, the performance of the VQA has shown successful performance~\cite{lu2019vilbert,tan2019lxmert}; however, it is not straightforward to apply the architectures to the video domain.
Compared to the image and text, video data involves more complex semantic structures along with not only spatial but also temporal-axis.
As described in Figure~\ref{fig:vqa_example}, long videos inevitably contain multiple events, and the events can have multiple and complex correlations.
Therefore, it is important to temporally ground the multiple events and their semantic structure.
\blfootnote{* indicates equal contribution\vspace{-0.8cm}}

Most of the previous datasets proposed for VideoQA consist of relatively short clips containing an event or a single action class, with relatively easy questions~\cite{jang-IJCV-2019,mario,lei2019tvqa,xu2016msr-vtt}.
For this reason, the understanding of the short clips can be sufficiently addressed with image-based architectures by selecting a few representative frames from the clips.
However, in the case of long videos having various events and complex relationships between the events, conventional architectures struggle to learn over large timescales.
For these cases, it is essential to address the temporal grounding of the various events by considering enough frames within videos.

In this paper, we suggest a novel video/text understanding method for intricate VideoQA tasks, which consist of complicated questions requiring multiple reasoning steps.
The two main ideas of the suggested methods are 1) efficiently sampling as many as possible informative visual features from the videos to learn the inherent temporal semantic structures and 2) considering the hierarchical dependency model to understand complex questions requiring multiple reasoning steps.

First of all, we suggest a deformable sampling module, which allows dense but efficient visual token sampling.
Obviously, the conventional sparse sampling method~\cite{lei2021less}, selecting a few frames followed by temporal pooling to get a single feature vector to be applied to downstream tasks, causes incomplete understanding for the long and intricate videos.
As can be seen in Figure~\ref{fig:vqa_example}, at least 3 intervals from the video should be considered to get a correct answer to the question.
Unfortunately, there is a fundamental trade-off between computational cost and the number of frames to be learned with the model.
To settle the problem, we introduce a deformable attention module that effectively selects a subset of meaningful visual features along spatio-temporal axis.
Specifically, the suggested method considers the semantics of the given query sentence.

Secondly, we introduce a dependency attention module to learn dependency-aware feature vectors of question tokens. 
As the input videos contain a more complex semantic structure, getting complicated questions requiring multiple reasoning steps are inevitable.
Therefore, it is necessary to take into account the semantic structure within questions to learn desirable embeddings of the question tokens.
We suggest leveraging the semantic structure from the dependency parse tree of the questions.
By combining the deformable sampling module and the dependency attention module,
our method is able to deal with the intricate compositional reasoning problems.

In experiments, we evaluate our model on Action Genome Question Answering (AGQA,\cite{GrundeMcLaughlin2021AGQA}) dataset.
The AGQA dataset is one of the most challenging benchmarks for VideoQA because it requires complex reasoning steps on long videos.
Extensive experiments not only show impressive quantitative results on QA accuracy but also verify the effectiveness of each module by a comprehensive ablation study.

In summary, our contributions are as follows:
\begin{itemize}
    \item We empirically reveal that covering a long time span is advantageous for complex problems, which needs spatio-temporal reasoning.
    \item We introduce a deformable sampling-based VideoQA model, DSR, which aims to solve compositional reasoning problem.
    \item Our experiments on VideoQA benchmarks show that the proposed method has the ability to perform complex spatio-temporal reasoning.
\end{itemize}

%------------------------------------------------------------------------
\section{Related Work}
\paragraph{Visual Question Answering}
VQA is the task of understanding how two inputs, text-based questions and visual features, relate to one another, proposed by Antol~\etal~\cite{Antol_2015_ICCV}.
For image-based question answering tasks, a significant amount of works propose attention-based model architectures to fuse question and image representations~\cite{Anderson2017up-down,zhou2019vlp,Kim2018,gao2019dynamic}.
Kim~\etal~\cite{Kim2018} show remarkable performance by utilizing a bilinear attention network that finds bilinear interactions between two modalities.
Moreover, inspired by the recent success of pre-trained language models~\cite{devlin-etal-2019-bert,clark2020electra}, universal pre-training frameworks for a vision-language representation learning achieve state-of-the-art performances not only on VQA but also on general visual-language tasks~\cite{tan2019lxmert,lu2019vilbert}.

However, question-answering in the video domain is under-explored compared to those in the image domain. 
Contrary to the growing interest in measuring video reasoning capabilities~\cite{jang-IJCV-2019,GrundeMcLaughlin2021AGQA,lei2018tvqa,lei2019tvqa,xu2017video}, existing VideoQA models mostly deal with short clip videos or simple questions~\cite{le2020hierarchical,fan-CVPR-2019,seo-etal-2021-attend,lei2021less}.
Since a video is a sequence of images containing the temporal dimension, understanding richer spatio-temporal features and temporal localization of natural language is essential.
To fuse the temporal feature, Fan~\etal~and Seo~\etal~attempt to utilize separate motion and appearance feature modules and integrate them with additional fusion network~\cite{le2020hierarchical,fan-CVPR-2019,seo-etal-2021-attend}.
Le~\etal~\cite{le2020hierarchical} propose a hierarchical conditional relation network to embed the video input at different granularities. 
However, separated modules have limitations in effectively interacting with linguistic questions, and the performances fell behind as the transformer-based model rises.
Kim~\etal~\cite{self-videoqa} propose a contrastive learning based training scheme that shows competitive performance, but only specializes in multiple-choice tasks.
The current state-of-the-art model in VideoQA is ClipBERT~\cite{lei2021less}, which is based on a cross-modal transformer.
ClipBERT enables end-to-end learning by employing sparse sampling while it is unsuitable for intricate tasks that require advanced spatio-temporal reasoning since random sparse sampling loses several semantic structures.
We propose a dense but efficient VideoQA model based on a transformer, which can maintain whole semantic structures.
\vspace{-0.7cm}

\paragraph{Efficient Transformers}
Transformer architecture~\cite{NIPS2017_3f5ee243} has shown remarkable performance on various downstream tasks. However, computational cost and memory consumption of Transformer increase quadratically depending on the length of input sequences.
There has been a surge of research interests recently in exploiting efficient transformer architectures to mitigate the problem.
For example, various algorithms that approximate the quadratic cost attention matrix based on low-rank matrix factorization have been proposed in the field of natural language processing~\cite{wang2020linformer,choromanski2021rethinking,xiong2021nystromformer}. 
Also, in the vision domain, the scope of self-attention is restricted to local neighborhoods or specific axis based on the locality assumption of objects~\cite{ho2019axial,child2019generating,gberta_2021_ICML,Arnab_2021_ICCV}.
However, the above algorithms are defined for the single modality.
In contrast, we aim to address cross-modal sparsification based on a question conditional visual token sampling algorithm for the VideoQA task where non-local and fine-grained features are required. 

%-------------------------------------------------------------------------
\section{DSR: Deformable Sampling-based VideoQA model for Compositional Reasoning}
In this section, we introduce a detailed explanation of our model. We consider the intricate VideoQA problems, which require compositional spatio-temporal reasoning. Our goal is to learn a generalizable visual-reasoning representation with deformable sampling and dependency modeling.

\begin{figure}
\begin{center}
\includegraphics[width=0.75\linewidth]{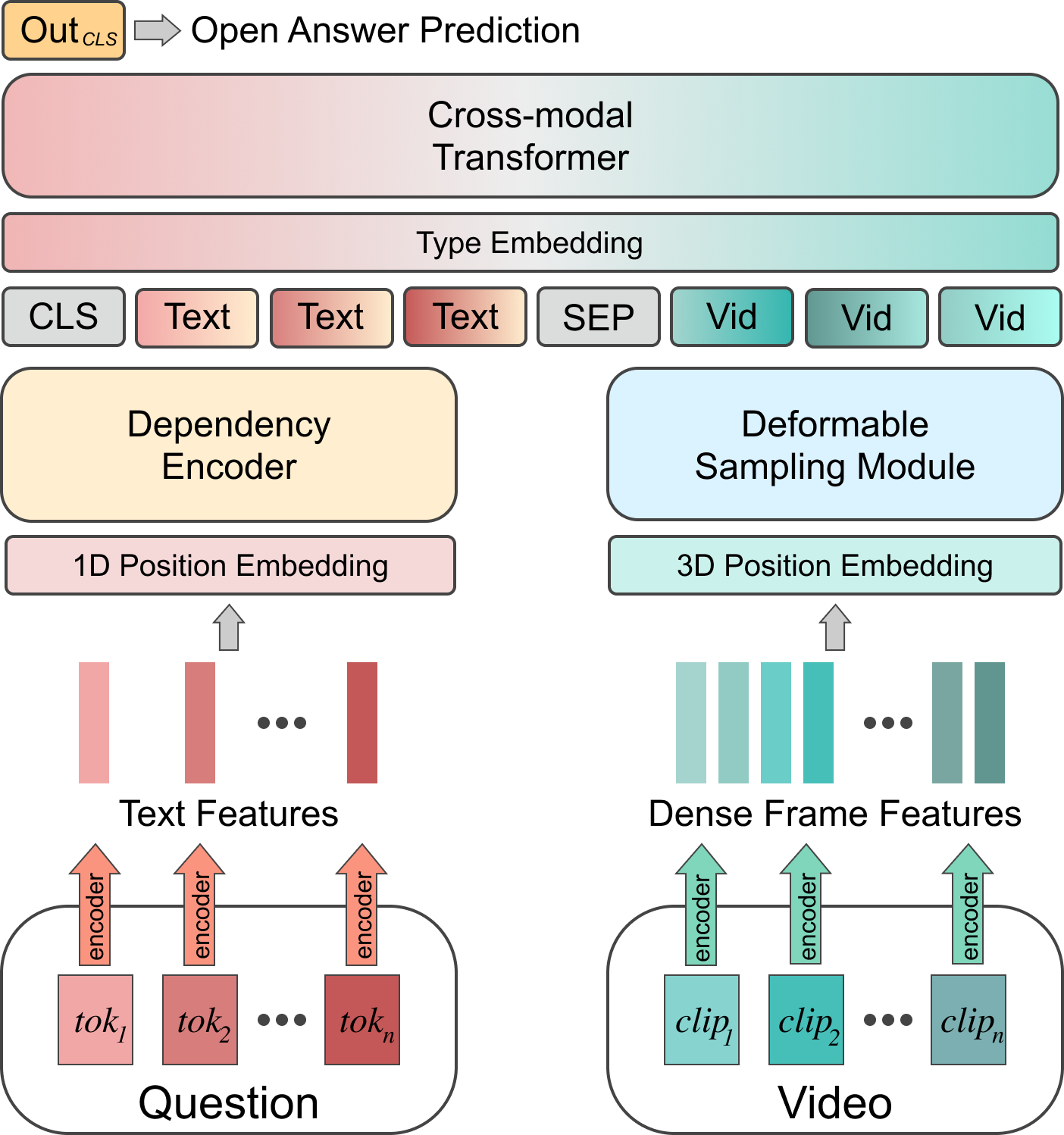}
\end{center}
% \vspace{-0.5cm}
   \caption{The overall architecture of DSR. Two details are missed for simplification; deformable sampling is conditioned on question context embedding, and global visual features are used as additional input to the cross-modal transformer.}
\label{fig:overall}
\vspace{-0.2cm}
\end{figure}

% \begin{figure}[t]
%     \centering
%     \begin{minipage}[t]{0.5\linewidth}
%         \centering
%         \includegraphics[width=\linewidth]{image/overall.png}
%         \caption{The overall architecture of DSR.}
%         \label{fig:overall}
%     \end{minipage}
%     \begin{minipage}[t]{0.45\linewidth}
%         \centering
%         \includegraphics[width=0.8\linewidth]{image/deform.png}
%         \caption{Illustration of the proposed deformable sampling module. The figure represents a single head of a single CDA layer. For simplicity, we only visualize a deformable attention procedure of one reference point, which is solely colored as blue.}
%         \label{fig:main_figure}    
%     \end{minipage}
%     % \vspace{-0.4cm}
% \end{figure}

\subsection{Transformer-based dense sampling model}
We propose Deformable Sampling-based VideoQA model for compositional Reasoning (DSR), dense but efficient one that utilizes deformable sampling for video features and dependency modeling for text questions.
Figure~\ref{fig:overall} gives an overall architecture of DSR, which is based on a cross-modal transformer. 
Each visual feature and question token are independently encoded with a vision backbone model and a language encoder, respectively.
Inputs of the cross-modal transformer are conditionally sampled video features and dependency guided question tokens.
We denote visual and language inputs of transformer as $V=[v_{1},v_{2},...,v_{L_v}]\in \mathbb{R}^{d\times L_v}$ and $L=[l_{1},l_{2},...,l_{L_q}]\in \mathbb{R}^{d\times L_q}$, respectively, where $L_v$ is the number of visual tokens sampled from conditional sampling module, $L_q$ represents the number of question tokens, and $d$ indicates representation dimension.
These embeddings of two different modalities are concatenated as input to a 12-layer transformer for cross-modal fusion, with special tokens [CLS] and [SEP].

We first uniformly sample the frames from a video, which is sufficiently dense to cover the full length of a video.
However, as the length of the video increases, usage of whole dense frames becomes impossible since it can not fit into a single transformer due to memory limitations.
Thus, motivated by Zhu~\etal~\cite{zhu2020deformable}, we introduce a deformable sampling module to only sample necessary visual features from full dense frames, conditionally to question embeddings.
Consequently, relatively few visual features compared to initial dense features are sampled from the module. 
A detailed explanation of conditional sampling is stated in Section~\ref{sec: deform}.
Language inputs (\textit{i.e.,} question tokens) also go through the pre-stage modeling step to enable compositional reasoning.
The dependency attention module forces specific attention head of transformer to understand dependency parsing structure, representing the relationship between words in a question sequence. It will be explained in Section~\ref{sec: dep}. 

The output vector of [CLS] token, $\mathbf{h_{cls}}$, is an aggregated representation of the entire input sequence of the cross-modal transformer, used to predict the answer.
We consider all the QAs as open-ended word tasks, which choose one correct word as the answer from a pre-defined answer set of size C.
We calculate a classification score by applying a linear classifier and softmax function on the final output and train the model by minimizing cross-entropy loss,
\begin{equation}
    \small
    L_{open}=-
    \sum_{c=1}^{C}\mathbbm{1}\{y=c\} \log(p_{c}),
\end{equation}
in which $\mathbf{p} = softmax(FFN(\mathbf{h_{cls}}))\in\mathrm{R}^{C}$ and $y$ is the ground truth answer label.
During inference, conditionally sampled visual features and dependency modeled linguistic features are utilized to predict answers with proper reasoning, in the same manner with the training phase.
In summary, our model achieves state-of-the-art performance on intricate VideoQA tasks by allowing end-to-end learning while covering temporally long and spatially fine-grained visual features, which are both important for advanced modeling.
Different from the model that only observes a single or a few video clips, DSR can tackle the data that needs compositional reasoning.

\subsection{Conditional Visual Feature Sampling}
\label{sec: deform}
In this section, we describe how to effectively sample a subset of visual tokens from the long and dense feature map.
Since video data has an additional temporal axis compared to image data, the feature map size of a video clip is much bigger than the feature map from an image. Thus, most of the VideoQA algorithms pool the feature map spatially~\cite{li2020hero} or temporally~\cite{lei2021less} and concatenate the sequence of visual features to question word vectors. Then, the concatenated feature is used as an input for a transformer-based QA model. 
However, the pooling-based approach would be sub-optimal for compositional VideoQA tasks that require long and fined-grained visual cues. 
Here, we assume that most visual features in spatio-temporal feature maps are redundant and uninformative for answering given questions. In the next section, we describe how to sample a few informative visual features from the dense feature map. 

\begin{figure}
\begin{center}
\includegraphics[width=0.65\linewidth]{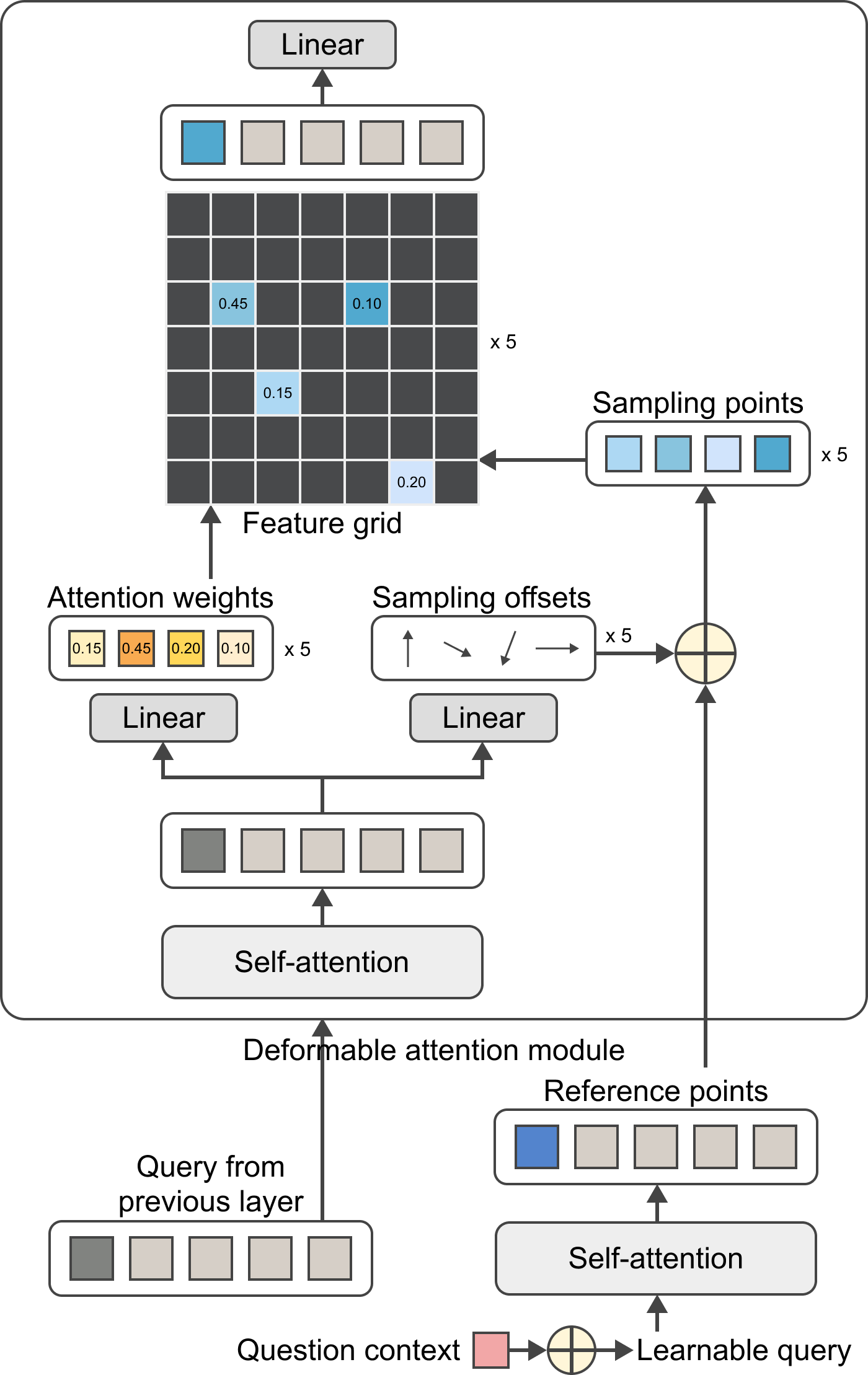}
\end{center}
% \vspace{-0.5cm}
\caption{Illustration of the proposed deformable sampling module. The figure represents a single head of a single CDA layer. For simplicity, we only visualize a deformable attention procedure of one reference point, which is solely colored as blue.}
\label{fig:main_figure}
\vspace{-0.3cm}
\end{figure}

\paragraph{Conditional Deformable Attention } 
Let $X \in{\mathbb{R}^{d \times t \times h \times w}}$ be a dense visual feature map extracted by a visual encoder such as ResNet~\cite{he2016deep}. The $d, t, h$, and $w$ indicate dimension, temporal length, height, and width of the feature map, respectively.  
Based on the 2-d deformable attention module~\cite{zhu2020deformable}, we define our 3-d Conditional Deformable Attention (CDA) to sample question-conditional visual features from the spatio-temporal feature map $X$ and a given question $L$ as follows:
\begin{equation}
    \label{eq:cda}
    \footnotesize
    \begin{split}
        \mathrm{CDA}(\mathbf{z}_q, p_q, X, L) =
        \sum_{m=1}^{M} W_{m}[\sum_{k=1}^{K} A_{mqk}\cdot W'_{m}X(p_q + \Delta p_{mqk})],\\
        where, A_{mqk} = W_{m}^{A} \bar{\mathbf{z}}_q, \quad \Delta p_{mqk} = W_{m}^{\Delta p} \bar{\mathbf{z}}_q, \quad \bar{\mathbf{z}}_q = \mathbf{z}_q \oplus pool(L), 
    \end{split}
\end{equation}
% \begin{equation}
%     \label{eq:cda}
%     \footnotesize
%     \begin{split}
%         \mathrm{CDA}(\mathbf{z}_q, p_q, X, L) &= \\
%         \sum_{m=1}^{M} W_{m}[\sum_{k=1}^{K} & A_{mqk}\cdot W'_{m}X(p_q + \Delta p_{mqk})],
%     \end{split}
% \end{equation}
in which $q$ is an element index for input query vector $\mathbf{z}_q$ of a transformer layer and 3-d reference point $p_q$. 
In the first transformer layer, the input query $\mathbf{z}_q \in \mathbb{R}^{L_v \times d}$ is the learnable query, where $L_v$ is the number of queries which is the same as the number of sampled visual features. Also, before feeding $\mathbf{z}_q$ to the first transformer layer, we make a pooled question context $\hat{L} = pool(L) \in \mathbb{R}^{1 \times d}$ and add the question context to each learnable query by the broadcast vector addition, $\oplus$, to make CDA sample visual features conditioned on the given question context.  
For the rest of the layers, the $\bar{\mathbf{z}}_q$ is the output vector of the previous transformer layer. $M$ and $K$ denote the total number of attention heads and sampled key vectors, respectively. $W$, $W'$, $W_{m}^{A}$, and $W_{m}^{\Delta p}$ are learnable linear projection layers. $A_{mqk}$ denotes the attention weight of the $k^{th}$ sampling point in the $m^{th}$ attention head for a given query $\bar{\mathbf{z}}_q$, where $\sum_{k=1}^{K}A_{mqk} = 1$. $\Delta p_{mqk} \in \mathbb{R}^3$ is 3-d sampling offset. Since $p_q + \Delta p_{mqk}$ is a real-valued vector, we apply trilinear interpolation to compute $X(p_q + \Delta p_{mqk})$.
%At the first CDA layer, reference points $p_q$ are obtained onetime via a linear projection over the learnable object queries. Then the reference points are just referred to the rest of the layers.
With CDA, we can get $L_v$ sampled visual tokens $V \in \mathbb{R}^{d \times L_v}$ where $L_v$ is much smaller than the $t \times h \times w$, \textit{e.g.,} 25 vs. 30 $\times$ 7 $\times$ 7.
The overview of CDA is illustrated in Figure~\ref{fig:main_figure}. 

\paragraph{Regularization for Sampling Diversity } 
For the question and answering task, sampled visual tokens from CDA are concatenated with question words, and a transformer-based model takes the concatenated features to predict an answer. 
Thus, it is important that the sampled visual tokens should be as diverse as possible to provide sufficient information for a given question. 
In Deformable DETR~\cite{zhu2020deformable}, offset predictions can be diverse without the collapse since each object query is trained to match a target object based on the Hungarian loss. 
However, in QA task, proper regularization is crucial to prevent the collapse because the model gets gradient feedback only from answering loss.
To reinforce the diversity of sampled visual tokens, we explore three types of additional regularization terms. Here, we consider batched features where the $V$ and $X$ have shapes of $(N \times d \times L_v)$ and $(N \times d \times thw)$, respectively.
The first regularization term is Soft Orthogonality (SO)~\cite{xie2017all}, which is defined as follows:
\begin{equation}
    \label{eq:so}
    \small
    \begin{split}
        \mathrm{(SO)} \qquad \lambda \sum_{i=1}^{N}\lVert V_i^{\top} V_i - I \rVert_{F}^{2},
    \end{split}
\end{equation}
where $i$ indicates the index in a mini-batch. The SO aims the Gram matrix of the sampled tokens to be close to the identity matrix, $I$. Thus, each sampled visual token could be distinctive and independent. 

The second regularization term is Maximal Coding Rate (MCR)~\cite{yu2020learning}, which formulated as follows:
\begin{equation}
    \label{eq:mcr}
    \small
    \begin{split}
        \mathrm{(MCR)} \qquad -\lambda \sum_{i=1}^{N}\frac{1}{2}\mathrm{log\ det}\left(I + \frac{d}{L_v\epsilon^2} V_i^{\top} V_i\right),
    \end{split}
\end{equation}
Maximizing the MCR results in the largest possible volume spanned by the vectors in the Gram matrix of the sampled tokens. Thus, the sampled tokens should be as independent as possible. 

The last regularization term that we explore is the contrastive loss~\cite{chen2020simple}. Here, we set the anchors, positive, and negative examples as sampled visual features $V_i$, the feature map $X_i$, and feature maps of others $X_{j \neq i}$ in the batch, respectively. 
\begin{equation}
    \label{eq:cont}
    \footnotesize
    \begin{split}
        \mathrm{(Contrastive)} \ -\lambda \sum_{i=1}^{N} \mathrm{log}\frac{\mathrm{exp(sim}(\hat{V}_i, \hat{X}_i) / \tau)}{\sum_{j=1}^{N}\mathbbm{1}_{[j \neq i]} \mathrm{exp(sim(} \hat{V}_i, \hat{X}_j) / \tau ) },
    \end{split}
\end{equation}
where $\hat{V}$ and $\hat{X}$ are the global averaged pooled  of $V$ and $X$, respectively. Also, we use the cosine similarity as the similarity function $\mathrm{sim}(\cdot, \cdot)$ and $\tau$ is set to 0.1 by default.

\paragraph{Global Context Features } 
The sampled visual features from our CDA represent fine-grained local information that is required to answer the given question. However, the interaction between local and global information is also crucial to solving the spatio-temporally complex QA task more accurately. Thus, we introduce additional global information $X_g \in \mathbb{R}^{N \times d \times t}$ that is extracted by applying the spatial pooling to feature maps $X \in \mathbb{R}^{N \times d \times t \times h \times w}$. As a result, we get the global-local visual feature $X_{gl} \in \mathbb{R}^{N \times d \times (L_v + t)}$ by concatenating the global and sampled local visual features. 

% \begin{figure}[t]
%     \centering
%     \begin{minipage}[t]{0.55\linewidth}
%         \centering
%         \includegraphics[width=\linewidth]{image/dep.png}
%         \caption{Example of dependency structure.}
%         \label{fig:depa}
%     \end{minipage}
%     \begin{minipage}[t]{0.4\linewidth}    
%         \centering
%         \includegraphics[width=0.7\linewidth]{image/dep_att.png}
%         \caption{Adjacency matrix generated from dependency relations.}
%         \label{fig:depb}
%     \end{minipage}
%     % \vspace{-0.3cm}
% \end{figure}
\begin{figure}
    \centering
    \includegraphics[width=\linewidth]{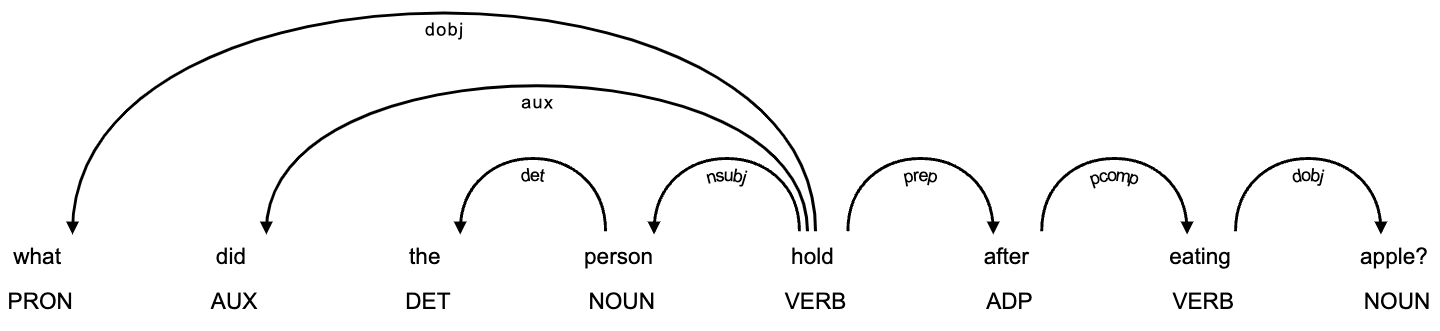}
    \caption{Example of dependency structure.}
    \label{fig:depa}
    % \vspace{0.2cm}
\end{figure}

\begin{figure}
    \centering
    \includegraphics[width=0.45\linewidth]{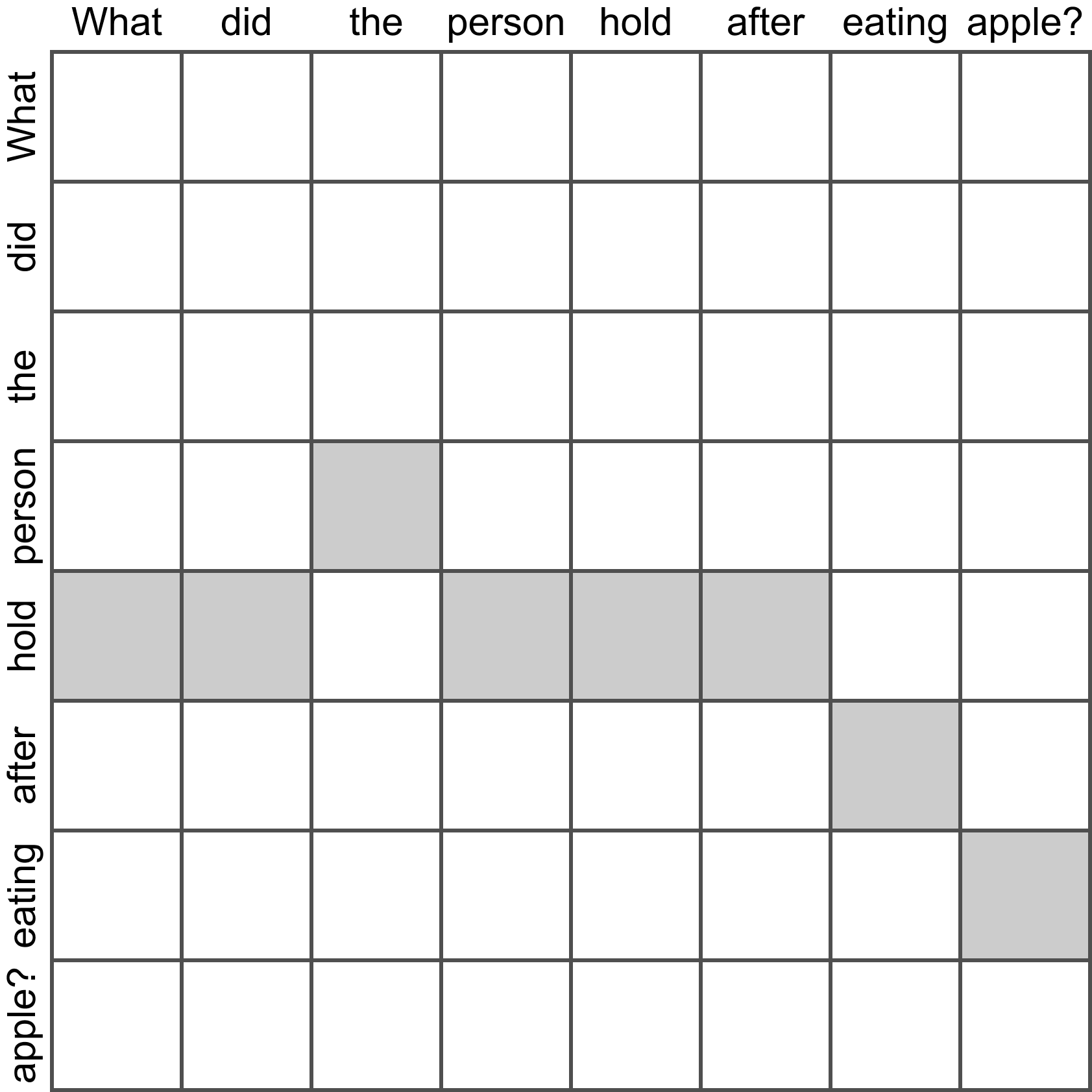}
    \caption{Adjacency matrix generated from dependency relations.}
    \label{fig:depb}
\vspace{-0.2cm}
\end{figure}

\subsection{Dependency Attention Module}
\label{sec: dep}
In this section, we explain the details of the dependency attention module that extracts dependency-aware vector from question tokens. Motivated from Deguchi~\etal~\cite{deguchi-etal-2019-dependency}, we introduce a self-attention module that incorporates dependency relations.
Previous studies show that the performance of neural machine translation has been improved by incorporating sentence structures~\cite{chen-etal-2017-neural,eriguchi-etal-2017-learning,wu}.
While most visual-language learning tasks only rely on a pre-trained language model to encode question embeddings, we believe that comprehension of sentence structure is crucial for nonconventional questions, and the dependency-based attention module would also work for the VideoQA task.

Language features are first learned through the dependency attention module before feeding into the cross-modal transformer.
The module is consist of an $L$-layer transformer, where one attention head of the $i$-th multi-head self-attention layer is trained with constraints based on dependency parsing value.
Let $O^{i-1} \in \mathbb{R}^{L_q \times d}$ is the output of previous layer.
The dependency attention module first maps $O^{i-1}$ to $d_{head}$-dimensional subspaces of multi-head attention as
\begin{equation}
    \small
    Q_{dep} = O^{i-1}W^{Q_{dep}},
\end{equation}
\begin{equation}
    \small
    K_{dep} = O^{i-1}W^{K_{dep}},
\end{equation}
\begin{equation}
    \small
    V_{dep} = O^{i-1}W^{V_{dep}},
\end{equation}
where $W^{Q_{dep}}$, $W^{K_{dep}}$, and $W^{V_{dep}}$ are $d \times d_{head}$ dimensional parameter matrices.
Dependency attention weight matrix, $A_{dep}$, is calculated by the bi-affine operation~\cite{DBLP:conf/iclr/DozatM17} as follows,
\begin{equation}
    \small
    A_{dep} = softmax(Q_{dep}UK^{T}_{dep}),
\end{equation}
where $U\in\mathbb{R}^{d_{head}\times d_{head}}$.
Each value of $A_{dep}$ represents the dependency relationship between two words, and the probability of token $q_{1}$ being the governor of token $q_{2}$ is modeled as $A_{dep}[q_{2},q_{1}]$.
Then, likewise the original self-attention module, attention output is obtained by multiplying $A_{dep}$ and $V_{dep}$.
% , $M_{dep} = A_{dep}V_{dep}$.
Finally, attention outputs of all the heads (\textit{i.e.,} one dependency outputs and $n_{head}-1$ conventional outputs) are concatenated and the rest are calculated like the conventional multi-head attention.

% \begin{figure}
% \vspace{-0.6cm}
% \centering
% \begin{subfigure}[t]{0.45\linewidth}
%     \centering
%     \includegraphics[width=\linewidth]{image/dep.png}
%     \subcaption{Example of dependency structure.}
%     \label{fig:depa}
% \end{subfigure}
% \begin{subfigure}[t]{0.45\linewidth}
%     \centering
%     \includegraphics[width=0.6\linewidth]{image/dep_att.png}
%     \subcaption{Adjacency matrix generated from dependency relations.}
%     \label{fig:depb}
% \end{subfigure}
% \vspace{-0.3cm}
%   \caption{Illustration of dependency attention module}
% \label{fig:dep}
% \vspace{-0.6cm}
% \end{figure}

Although $A_{dep}$ can be learned by additional dependency loss function as in Deguchi~\etal~\cite{deguchi-etal-2019-dependency}, we explicitly force the correct dependency value not only in the training phase but also in inference.
The gold parse provides an upper bound for using dependency relations and enables accurate structural modeling.
Figure~\ref{fig:depa} shows the example of dependency relationships and Figure~\ref{fig:depb} represents the relations in an adjacency matrix that utilized as a gold value of $A_{dep}$.
The gold value forces each token to only give attention to its governor.

To apply dependency relations to the transformer module, we reorganize the adjacency matrix for subword sequences.
When a word is separated into multiple subwords by BPE~\cite{sennrich-etal-2016-neural}, the governor (\textit{i.e.,} the head) of the rightmost subword is set to the governor of the original word and the governor of each subword other than the rightmost one is set to the right adjacent subword.

%-------------------------------------------------------------------------
\section{Experiment}
In this section, we evaluate our proposed model on compositional spatio-temporal reasoning datasets. 
We first introduce the details of benchmark datasets in Section~\ref{sec: dataset}. Section~\ref{sec: setup} describes experimental setup including implementation details.
We also provide extensive quantitative experiments and ablation studies in Section~\ref{sec: quantitative}, to show how each of the proposed modules works.
Lastly, we qualitatively confirm that our model samples reasonable visual frames conditioned on given questions, in supplementary materials.

\subsection{Dataset}
\label{sec: dataset}
\paragraph{Action Genome Question Answering}
We validate DSR using AGQA dataset proposed by Grunde-McLaughlin~\etal~\cite{GrundeMcLaughlin2021AGQA}, the most challenging benchmark for VideoQA.
While most of the existing benchmarks only utilize short video clips, use simple and biased questions, and focus on questions that require commonsense or external knowledge, AGQA consists of long video clips with an average length of 30 seconds. 
Each question generated by a handcrafted program necessarily requires spatio-temporal reasoning steps.

We adopt a balanced, novel composition, and more composition version of AGQA~\cite{GrundeMcLaughlin2021AGQA}.
A balanced dataset, 3.9M of QA pairs associated with 9.6K videos, minimizes the bias by balancing the answer distributions and types of question structures.
Novel composition is constructed to test whether models can disentangle distinct concepts and combine them well. For example, compositions like ``before standing up'' are removed from the training set, while each word ``before'' and ``standing'' appear.
It tests how well the model performs on questions with those novel compositions in inference.
More composition tests whether models generalize to more compositional steps.
The training set only contains simpler questions with $\leq S$ compositional steps, while the test set contains only questions with $> S$ reasoning steps.
The model generalized to novel compositions and more compositional steps can be regarded as a successful VideoQA model that understands compositional semantic structures.

Open answer questions have many possible answers, while binary questions have answers that are yes/no or before/after.
Except for the first table, all the tables adopt a 10\% version of the balanced dataset for the training and inference phase.
We provide details of the dataset we used in supplementary materials.
\vspace{-0.3cm}

\paragraph{TVQA}~\cite{lei2018tvqa} is an intricate multiple-choice VideoQA dataset composed of 60-90 second long video clips.
Although the video clips, questions, answers, subtitles, timestamps, and objects are given from the dataset, we only utilize video clips and QA pairs to verify intricate compositional reasoning ability.
Most baselines use subtitles and propose the model that maximizes the subtitle knowledge since the performance gained from subtitles is much larger than that from video, however, we claim only to use videos, questions, and answers for \textit{video question answering} tasks to show the video understanding ability.
\vspace{-0.3cm}

% \paragraph{MSRVTT-QA} is one of the most famous benchmarks of VideoQA. 
\paragraph{Benchmarks with short videos} 
MSRVTT-QA~\cite{xu2016msr-vtt} is created based on videos in MSRVTT and questions are automatically generated from video descriptions. It consists of 10k videos and 243k QA pairs, with an average video length of 15 seconds.
TGIF-QA~\cite{jang2017tgif} is web GIF VQA, containing 165K QA pairs on 72K GIF videos with an average length of 3 seconds long.
MSRVTT and TGIF are not only short but also easy videos. These videos only require simple spatial reasoning while AGQA requires intricate spatio-temporal reasoning.
According to the original paper, MSRVTT is a set of simple clips that each can be described with a single sentence, thus confined to a single domain. 
Lei~\etal~\cite{lei2021less} support it by showing that adding more clips does not improve performance for both datasets; even somewhat has a negative effect.
Our model does not stand out in simple tasks since it aims to solve intricate reasoning problems by modeling dense features, but even shows competitive results on the datasets.

% As shown in Table~\ref{tab:tgifqa}, our DSR achieves better accuracy on $Action$ and $Transition$ tasks than the reproduced ClipBERT. Since the $Action$ and $Transition$ tasks require a model to understand long-term temporal contexts than the $FrameQA$ task\footnote{Example questions for the 3 tasks of TGIF-QA. $Action$: What does the cat do 3 times?, $Transition$: What does the model do after lower coat?, $FrameQA$: What is the color of the bulldog?.}, our DSR can show better performance in the two tasks. 

\begin{table*}
% \vspace{-0.3cm}
\setlength\tabcolsep{2.3pt}
\begin{center}
\scalebox{0.8}{
\begin{tabular}{lcccccc}
\midrule
& Types & PSAC~\cite{li2019beyond} & HME~\cite{fan-CVPR-2019} & HCRN~\cite{le2020hierarchical} & ClipBERT~\cite{lei2021less} & DSR(Ours)\\
\midrule
\multicolumn{1}{l}{\multirow{3}{*}{\parbox{2cm}{Full\phantom{abcde} Balanced}}} & 
\multicolumn{1}{c}{Binary} & \multicolumn{1}{c}{54.19} & \multicolumn{1}{c}{59.77} & \multicolumn{1}{c}{58.11} & \multicolumn{1}{c}{63.83} & \multicolumn{1}{c}{\textbf{65.92}}(+2.09) \\
& \multicolumn{1}{c}{Open} & \multicolumn{1}{c}{27.20} & \multicolumn{1}{c}{36.23} & \multicolumn{1}{c}{37.18} & \multicolumn{1}{c}{48.54} & \multicolumn{1}{c}{\textbf{49.54}}(+1.00) \\
& \multicolumn{1}{c}{All} & \multicolumn{1}{c}{40.40} & \multicolumn{1}{c}{47.74} & \multicolumn{1}{c}{47.42} & \multicolumn{1}{c}{53.03} & \multicolumn{1}{c}{\textbf{54.36}}(+1.33) \\
\midrule
\multicolumn{1}{l}{\multirow{3}{*}{\parbox{2cm}{Novel\phantom{abcde} Composition}}} &
\multicolumn{1}{c}{Binary} & \multicolumn{1}{c}{43.00} & \multicolumn{1}{c}{52.39} & \multicolumn{1}{c}{43.40} & \multicolumn{1}{c}{53.87} & \multicolumn{1}{c}{\textbf{59.57}}(+5.70) \\
& \multicolumn{1}{c}{Open} & \multicolumn{1}{c}{14.80} & \multicolumn{1}{c}{19.46} & \multicolumn{1}{c}{23.72} & \multicolumn{1}{c}{36.45} & \multicolumn{1}{c}{\textbf{38.73}}(+2.28) \\
& \multicolumn{1}{c}{All} & \multicolumn{1}{c}{32.49} & \multicolumn{1}{c}{40.11} & \multicolumn{1}{c}{36.06} & \multicolumn{1}{c}{40.82} & \multicolumn{1}{c}{\textbf{43.96}}(+3.14) \\
\midrule
\multicolumn{1}{l}{\multirow{3}{*}{\parbox{2cm}{More\phantom{abcde} Composition}}} & 
\multicolumn{1}{c}{Binary} & \multicolumn{1}{c}{35.39} & \multicolumn{1}{c}{\textbf{48.09}} & \multicolumn{1}{c}{42.46} & \multicolumn{1}{c}{42.93} & \multicolumn{1}{c}{47.79}(-0.30) \\
& \multicolumn{1}{c}{Open} & \multicolumn{1}{c}{28.00} & \multicolumn{1}{c}{33.47} & \multicolumn{1}{c}{34.81} & \multicolumn{1}{c}{45.93} & \multicolumn{1}{c}{\textbf{48.08}}(+2.15) \\
& \multicolumn{1}{c}{All} & \multicolumn{1}{c}{31.13} & \multicolumn{1}{c}{39.70} & \multicolumn{1}{c}{38.00} & \multicolumn{1}{c}{45.32} & \multicolumn{1}{c}{\textbf{48.02}}(+2.70) \\
\bottomrule
\end{tabular}
}
\end{center}
\caption{Quantitative comparison with the baselines on AGQA datset. Full balanced, novel composition, and more composition represent different subset of AGQA as described in Section~\ref{sec: dataset}. The bold represents the best score. 
% We run three independent trials and confirmed the statistical significance of DSR via t-test.
}
\label{tab:main_result}
% \vspace{-1.0cm}
\end{table*}

\subsection{Experimental Setup}
\label{sec: setup}
\paragraph{Baselines}
We compare our approach against four recent VideoQA methods~\cite{li2019beyond,fan-CVPR-2019,le2020hierarchical,lei2021less}.
PSAC~\cite{li2019beyond} utilizes a co-attention block after unimodal self-attention blocks to simultaneously attend to both modalities.
HME~\cite{fan-CVPR-2019} models question, appearance, and motion features with different LSTM encoders. Additional visual and question memories help the multimodal fusion.
HCRN~\cite{le2020hierarchical} designs conditional relation networks and stacks them to accommodate diverse input modalities and conditioning features.
ClipBERT~\cite{lei2021less} inputs a few short clips independently to a cross-modality transformer and aggregates prediction scores from multiple clips as the final score.
For PSAC, HME, and HCRN, the performances reported in Grunde-McLaughlin~\etal~\cite{GrundeMcLaughlin2021AGQA} are utilized.
\vspace{-0.3cm}

\paragraph{Implementation Details}
2D ResNet-50~\cite{resnet} and word embedding layers of BERT-base model~\cite{devlin-etal-2019-bert} are adopted as visual and language backbones.
Specifically, 5 Conv blocks of ResNet-50 and an extra convolution layer are used for spatial down-sampling. 
We initialize visual/text encoder and cross-modal transformer with image-text pretrained weights proposed from ClipBERT~\cite{lei2021less}, which leverages large-scale image-text datasets~\cite{coco,Krishna2016VisualGC}.
% (COCO Captions and Visual Genome Captions). 

We use a 4-layer transformer to construct our CDA. In each transformer layer, we set the number of attention heads and sampling points to 4 and 8, respectively.
Also, we use a 2-layer transformer for the dependency attention encoder. The first attention head of the first layer corresponds to a dependency-guided self-attention module.
3D and 1D positional embeddings are applied for visual and language embeddings, respectively.
We also add different type embeddings to both video and text inputs of the cross-modal transformer to indicate their source type. 
We report more details such as hyperparameters in supplementary materials.
% Input frames are resized to have a maximum longer side of L while keeping the aspect ratios, and the shorter side is zero-padded to be L as well [48]. 
% These weights are trained separately for their individual single-modality tasks, thus simply combining them together in a single framework for downstream task training may result in suboptimal performance. 

\subsection{Quantitative Results}
\label{sec: quantitative}
\paragraph{Comparision with baselines on varied benchmarks}
We compare DSR with state-of-the-art models on the aforementioned datasets.
As shown in Table~\ref{tab:main_result}, DSR consistently outperforms all the baselines on AGQA dataset.
Compared to the best baseline, ClipBERT, our model achieves better points of 1.33, 3.14, and 2.70 on full balanced, novel composition, and more composition datasets, respectively.
We run three independent trials on Table~\ref{tab:main_result},\ref{tab:msrvtt},\ref{tab: ablation} and confirmed the statistical significance of DSR via t-test.
The noticeable thing is that DSR, the model concentrates on complex spatio-reasoning, especially records high scores on novel composition and more composition subsets.
Since these two subsets are intentionally curated to test the generalizability and the reasoning ability of the model, the results give proof of the quality of DSR.
The experimental results according to the number of compositional steps are in supplementary materials.

% sets seemingly resemble FrameQA in TGIF-QA,
% as they tend to focus on spatial appearance features.
% This means that MASN is able to capture spatial
% details well

Table~\ref{tab:msrvtt} shows the results on MSRVTT-QA, TGIF-QA, and TVQA dataset. We experiment over three tasks (\textit{i.e.,} \textit{Action, Transition, FrameQA}) in TGIF-QA benchmark.
Even though the MSRVTT-QA and TGIF-QA mostly only require an understanding of the spatial features rather than temporal reasoning of given questions, our method achieves a score comparable with ClipBERT.
% Since the \textit{Action} and \textit{Transition} tasks require a model to understand long-term temporal contexts than the \textit{FrameQA}, DSR can show better performance in the two tasks.
% DSR and ClipBERT evaluated on TVQA, where an average clip length is 76s, achieve accuracy of 48.84 and 44.46, respectively. 
Moreover, DSR achieves the state-of-the-art result in the V+Q setting of TVQA, where subtitles and timestamps are not used for training.

\begin{table}
% \vspace{-0.3cm}
\setlength\tabcolsep{2.3pt}
\begin{center}
\scalebox{0.8}{
\begin{tabular}{lccccc}
\toprule
Methods & MSRVTT-QA & Action & Transition & FrameQA & TVQA\\
\midrule
Co-Memory~\cite{DBLP:conf/cvpr/GaoGCN18} & 32.0 & 68.2 & 74.3 & 51.5 & -\\
PSAC~\cite{li2019beyond} & - & 70.4 & 76.9 & 55.7 & -\\
HME~\cite{fan-CVPR-2019} & 33.0 & 73.9 & 77.8 & 53.8 & -\\
HCRN~\cite{le2020hierarchical} & 35.6 & 75.0 & 81.4 & 55.9 & -\\
QueST~\cite{jiang2020divide} & 34.6 & 79.5 & 81.0 & \textbf{59.7} & - \\
% MASN~\cite{seo-etal-2021-attend} & 35.2 & \\
multi-stream~\cite{lei2018tvqa} & - & - & - & - & 43.8\\
ClipBERT~\cite{lei2021less} & \textbf{37.4} & \textbf{82.4} & 87.3 & 58.8 & 44.4\\
DSR(Ours) & 37.2 & 81.7 & \textbf{87.6} & 58.3 & \textbf{48.8}\\
\bottomrule
\end{tabular}
}
\end{center}
% \vspace{-0.9cm}
\caption{Experimental results on benchmark datasets. 
% We run three independent trials and confirmed the statistical significance of DSR via t-test.
}
\label{tab:msrvtt}
\end{table}
%clipbert 우리가 구현한 것 언급

% \begin{table}
% \setlength\tabcolsep{4.5pt}
% \begin{center}
% \begin{tabular}{l|ccc}
% \toprule
% Methods & Action & Transition & FrameQA \\
% \midrule
% Co-Memory~\cite{DBLP:conf/cvpr/GaoGCN18} & 68.2 & 74.3 & 51.5 \\
% PSAC~\cite{li2019beyond} & 70.4 & 76.9 & 55.7 \\
% HME~\cite{fan-CVPR-2019} & 73.9 & 77.8 & 53.8 \\
% HCRN~\cite{le2020hierarchical} & 75.0 & 81.4 & 55.9 \\
% QueST~\cite{jiang2020divide} & 79.5 & 81.0 & \underline{59.7} \\
% ClipBERT$^{\dagger}$~\cite{lei2021less} & 82.0 & 87.3 & 58.7 \\
% DSR(Ours) & \underline{82.6} & \underline{87.4} & 58.1 \\
% \bottomrule
% \end{tabular}
% \end{center}
% \vspace{-0.5cm}
% \caption{Experimental results on TGIF-QA dataset. $^{\dagger}$ indicates the reproduced result with the same configurations described in the ClipBERT\cite{lei2021less} paper. The bold represents the best score and the second best is underlined.}
% \label{tab:tgifqa}
% \vspace{-0.1cm}
% \end{table}

% \begin{table}
% \setlength\tabcolsep{4.5pt}
% \begin{center}
% \caption{Experimental results on MSRVTT-QA dataset. The bold represents the best score and second best is underlined.}
% \label{tab:msrvtt}
% \begin{tabular}{lc}
% \toprule
% Methods & MSRVTT-QA Acc.\\
% \midrule
% Co-Memory~\cite{DBLP:conf/cvpr/GaoGCN18} & 32.0 \\
% HME~\cite{fan-CVPR-2019} & 33.0 \\
% HCRN~\cite{le2020hierarchical} & 35.6 \\
% MASN~\cite{seo-etal-2021-attend} & 35.2 \\
% ClipBERT~\cite{lei2021less} & \textbf{37.4} \\
% DSR(Ours) & \underline{37.2} \\
% \bottomrule
% \end{tabular}
% \end{center}
% \vspace{-0.6cm}
% \end{table}

\paragraph{QA Performance based on Sequence Length of Visual Features } 
Here, we analyze the efficiency and effectiveness of DSR when addressing long sequence visual features. We first explore how the QA accuracy varies as we increase the number of frames so that the visual feature covers a longer temporal range. In this experiment, we set fps as 1 by default.
From Table~\ref{tab:dense}, we observe that the QA accuracy increases as we show more frames to the model.
However, since the computational cost of self-attention operations in the transformer-based QA module increases quadratically based on the input sequence length, there is a limitation to consider a more long-ranged sequence without any sparsification of visual features. 
On the contrary, DSR can sample a subset of informative visual features from the dense feature map, the number of visual features can be controllable as a hyperparameter. As a result, we achieve higher accuracy even with a much less number of visual tokens (57 vs. 392). More detailed analyses of memory efficiency according to sequence lengths of our DSR are discussed in supplementary materials. 

%considering the  of self-attention operation of the transformer-based QA module, 
% frame수를 늘릴떄마다 v_token수도 같이 늘어남. QA module의 self-attention operation을 고려할 때 sequence-length 에 quadratic computation이 필요한데, 16 frames에서 한계 봉착
% 반면 DSR 사용시 더 긴 구간의 video를 보면서도 controllable 한 n_vtokens로 모델링 가능. 성능도 더 좋아

% \begin{figure}[t]
%     \centering
%     \begin{minipage}[t]{0.5\linewidth}
%         \centering
%         \includegraphics[width=\linewidth]{image/overall.png}
%         \caption{The overall architecture of DSR.}
%         \label{fig:overall}
%     \end{minipage}
%     \begin{minipage}[t]{0.45\linewidth}
%         \centering
%         \includegraphics[width=0.8\linewidth]{image/deform.png}
%         \caption{Illustration of the proposed deformable sampling module. The figure represents a single head of a single CDA layer. For simplicity, we only visualize a deformable attention procedure of one reference point, which is solely colored as blue.}
%         \label{fig:main_figure}    
%     \end{minipage}
%     \vspace{-0.6cm}
% \end{figure}

\begin{table}
    \begin{center}
        \scalebox{0.8}{
        \setlength\tabcolsep{4pt}
        \begin{tabular}{llccc}
        \toprule
        $N_{Frames}$ & $N_{V\_Tokens}$ & Binary & Open & All \\
        \midrule
        % 2 & 2$\times$7$\times$7 & 63.26 & 42.68 & 48.46 \\
        % 4 & 4$\times$7$\times$7 & 63.20 & 44.05 & 49.67 \\
        % 8 & 8$\times$7$\times$7 & 60.29 & 45.09 & 49.55 \\
        % 16 & 16$\times$7$\times$7 & 63.02 & 44.67 & 50.06 \\
        % 32 w/ DSR, 1fps & 32 + 25 & 60.17 & 46.56 & 50.56 \\
        2 & 2$\times$7$\times$7 & 60.22 & 46.05 & 50.21 \\
        4 & 4$\times$7$\times$7 & 60.32 & 47.40 & 51.20 \\
        8 & 8$\times$7$\times$7 & 61.29 & 46.37 & 50.75 \\
        % 16 & 16$\times$7$\times$7 & 63.42 & 48.94 & 53.19 \\
        32 w/ DSR & 32 + 25 & \textbf{64.47} & \textbf{48.58} & \textbf{53.24} \\
        %1fps 언급하기
        \bottomrule
        \end{tabular}}
    \end{center}
        \caption{Accuracy based on various sequence length of visual features. 
        % Our DSR achieves the highest accuracy with much fewer visual features compared to the model with a dense feature map. 
        }
        \label{tab:dense}
    \vspace{-0.3cm}
\end{table}

\paragraph{Sparse Sampling vs. Dense Sampling} 
In this experiment, we compare the effectiveness of randomly sampled sparse features and densely sampled features for the intricate compositional reasoning task. 
In the ClipBERT~\cite{lei2021less}, they propose a sparse sampling-based training strategy due to the high computation cost and memory consumption. 
The Sparse Random in Table~\ref{tab:sparse} follows the training convention of ClipBERT. They randomly sample multiple clips across the whole video, and each clip consists of 2 consecutive frames with fps 2. 
Then, a shared transformer-based QA model independently predicts answers based on the multiple clips. Finally, the answer logits from each clip are averaged as a final decision.
In contrast to sparse sampling, dense sampling aims to see longer sequences temporally with just one clip to address the intricate spatio-temporal reasoning task, AGQA.
% We also set the fps of dense sampling to 1; thus, a clip covers the 8 or 16 seconds range of a video.
We observe that dense sampling with DSR shows higher accuracy than sparse sampling. Since DSR can sample a few diverse informative visual features from the spatio-temporally dense feature map, the model can effectively associate question words and sampled visual features, which leads to the highest accuracy in Table~\ref{tab:sparse}.

\begin{table}
    \begin{center}
        \scalebox{0.8}{
        \setlength\tabcolsep{4pt}
        \begin{tabular}{lcccc}
        \toprule
        Sampling Method& $N_{Frames}$ & $N_{Clips}$ & Acc.\\
        \midrule
        \multicolumn{1}{l}{\multirow{3}{*}{Sparse Random}} & 
        \multicolumn{1}{c}{2} & \multicolumn{1}{c}{1} & \multicolumn{1}{c}{50.57} \\
        & \multicolumn{1}{c}{2} & \multicolumn{1}{c}{2} & \multicolumn{1}{c}{52.17} \\
        & \multicolumn{1}{c}{2} & \multicolumn{1}{c}{4} & \multicolumn{1}{c}{52.80} \\
        & \multicolumn{1}{c}{2} & \multicolumn{1}{c}{16} & \multicolumn{1}{c}{52.93} \\
        % Dense & 16 & 1 & 50.06 & \\
        % Dense & 16 & 1 & 53.19 & \\
        Dense w/ DSR & 32 & 1 & \textbf{53.24} & \\
        \bottomrule
        \end{tabular}}
    \end{center}
        \caption{Comparison of sparse sampling and dense sampling strategies.
        % While the pure dense sampling strategy show lower accuracy than the sparse sampling, dense sampling with DSR achieves the high accuracy.
        }
        \label{tab:sparse}
    \vspace{-0.3cm}
\end{table}

\vspace{-0.3cm}
\paragraph{Ablation study}
% 1 32 1 dep_layer 2
In this section, we conduct extensive ablation experiments about the hyperparameters of DSR.
The first row in Table~\ref{tab: ablation} is our best configuration among all controllable variables.
Firstly, we observe that there is an improvement with the dependency attention module. 
The dependency encoder helps structurization of question sequences by forcing dependency relations.
Then, we explore the best number of visual features to be sampled. When we increase the number of object queries from 5 to 25, there are consistent improvements in the QA accuracy. However, if we set the number of object queries to 50, the QA accuracy drops slightly. We analyze that noisy and redundant visual features could be sampled if we consider too many sampling points. Thus, we set the number of object queries to 25 by default for the experiments in prior sections. 

The next ablation is about the effectiveness of the global context features. From the sixth and seventh rows in Table~\ref{tab: ablation}, we observe that the global context features are notably helpful to boost the QA accuracy.
While the ``only local" model shows
% using only sampled local features shows the 
lower accuracy than the ``only global" model, we achieve the best performance with the combination of global features and local features. This indicates that a proper association of global and local features is crucial to address the complex spatio-temporal reasoning task. 

\begin{table}[t]
\begin{center}
\scalebox{0.8}{
\setlength\tabcolsep{4.5pt}
\begin{tabular}{c|c|c|c|c|c}
\toprule
Dep. & \# of Obj. q & Glob. & Reg. & Sampl. & Acc.\\
\midrule
\boldcheckmark & 32 + 25 & both & SO & Deform & \textbf{53.24} \\
\midrule
\boldxmark & 32 + 25 & both & SO & Deform & 52.26 \\
\midrule
\boldcheckmark & 32 + 5 & both & SO & Deform & 47.55 \\
\boldcheckmark & 32 + 10 & both & SO & Deform & 51.29 \\
\boldcheckmark & 32 + 50 & both & SO & Deform & 52.04 \\
\midrule
\boldcheckmark & 32 & only global & SO & Deform & 50.64 \\
\boldcheckmark & 25 & only local & SO & Deform & 45.72 \\
\midrule
\boldcheckmark & 32 + 25 & both & - & Deform & 50.97 \\
\boldcheckmark & 32 + 25 & both & Cont. & Deform & 49.84 \\
\boldcheckmark & 32 + 25 & both & MCR & Deform & 51.06 \\
\midrule
\boldcheckmark & 57 & local & - & Rand & 51.81 \\
% \boldcheckmark & 200 & local & - & Rand & 52.2 \\
\bottomrule
\end{tabular}
}
\end{center}
\caption{Results on ablation experiments.
% We run three independent trials and confirmed the statistical significance of DSR via t-test.
}
\label{tab: ablation}
\vspace{-0.5cm}
\end{table}

Subsequently, we explore the 3 types of sampling regularization terms. We find that the Soft Orthogonality (Eq~\ref{eq:so}) regularization achieves the best performance. The MCR regularization shows a high variance in the norm of gradients, which causes an unstable training process. We conjecture that the high variance comes from the $\mathrm{log det}$ operator. 
Also, the contrastive loss shows the lowest accuracy. This could be due to the small batch size caused by the 12-layer transformer and ResNet-50 taking video data as input. 

% In Deformable DETR, offset predictions can be diverse without the collapse since each object query is trained to match a target object based on the Hungarian loss. 
% However, in QA task, proper regularization is crucial to prevent the collapse because the model gets gradient feedback only from answering loss.
% Thus applying an effective regularizer is one of our main contributions, which elevates the performance of DSR by 2.3 points than that of DSR w/o regularizer.
Finally, we compare DSR to a random sampling strategy of visual features. For the random sampling strategy, we randomly sample 57 visual features from the dense feature map during the training phase. Then, we uniformly sample the visual features from the flattened dense feature map at the inference phase. 
% Since the feature maps come from a pre-trained ResNet50, each visual token can have meaningful semantic information. Thus, the performance drop was not significant with the random sampling strategy. 
As expected, DSR with the diversity regularization and the global-local fusion achieves higher accuracy by a large margin than the random sampling under the same number of visual tokens. 
This indicates that our two strategies, avoiding collapsed sampling and global-local information interaction, are essential in this sampling-based VideoQA task.

In addition, 
we 1) analyze the diversity and suitability of sampled tokens by visualizing the output of the deformable sampler, 2) visualize the effectiveness of the dependency attention module, and 3) validate that the performance gain of DSR comes from novel modules we proposed, but not from the increased parameter, in supplementary materials.

% \begin{table}
% \begin{center}
% \setlength\tabcolsep{4.5pt}
% \begin{tabular}{lcccc}
% \toprule
% Sampling Method& $N_{Frames}$ & $N_{Clips}$ & Acc.\\
% \midrule
% \multicolumn{1}{l}{\multirow{3}{*}{Sparse Random}} & 
% \multicolumn{1}{c}{2} & \multicolumn{1}{c}{1} & \multicolumn{1}{c}{50.57} \\
% & \multicolumn{1}{c}{2} & \multicolumn{1}{c}{2} & \multicolumn{1}{c}{52.17} \\
% & \multicolumn{1}{c}{2} & \multicolumn{1}{c}{4} & \multicolumn{1}{c}{52.80} \\
% Dense & 16 & 1 & 50.06 & \\
% Dense w/ DSR & 32 & 1 & \textbf{53.24} & \\
% \bottomrule
% \end{tabular}
% \end{center}
% \vspace{-0.5cm}
% \caption{}
% \label{tab:sparse}
% \vspace{-0.5cm}
% \end{table}

% reg object_query dependency sampling Acc multiscale

% \subsection{Visual Analysis}
% \label{sec: qualitative}
% Deformable sampling의 핵심은 dense한 feature 중에서 question 및 video의 형태에 conditional 하게 필요한 소수의 frame/feature 만 뽑아오는 것이다.
% 모듈의 output (transformer의 input)을 visualize 해서 얼마나 diverse하게 뽑아오는지, question에 필요한 부분만 뽑아오는지를 보여준다.
% 예시 2개 full column으로 보여주기.
% 추가적인 viz와, dependency attention module에 대한 visual analysis를 supplemantary materials.

% question query token에 대해 어떤 frame 뽑아오는지 attention 및 이미지 viz
% question given이 아닐때는 다양한걸 뽑아온다는걸 보여줘야 함

% dependency module 사용한거, 아닌거 각각 더 sparse하게 구조화 잘 된다는것 찾아보기

% \begin{figure}
% \begin{center}
% \includegraphics[width=\linewidth]{latex/image/graph.png}
% \end{center}
%     \vspace{-0.5cm}
%   \caption{flops vs performance}
% \label{fig:robust}
% \vspace{-0.5cm}
% \end{figure}

\section{Conclusion}
This paper presents the state-of-the-art compositional reasoning model for video question answering tasks, DSR, which utilizes deformable sampling module and dependency attention module for efficient video-text representation learning.
Based on our finding that the dense model performs better than the sparse model on the compositional reasoning dataset, which is a different point of view from previous work, we conditionally sample question-related visual features from a dense feature map.
This process remarkably reduces the number of visual tokens needed for cross-modal transformer while rather improving the efficiency; maximum allowed batch size and performance increase.
The dependency-based attention module eases the model to conduct multi-step reasoning by guiding a particular attention head with structuralized dependency relations.
Extensive experiments verify our model especially stands out against others on intricate benchmarks.
Comprehensive ablation studies demonstrate each factor fairly contributes to our model.

\vspace{0.2cm}
\small\noindent{\small\textbf{Acknowledgements}} 
This work was supported by Korea research grant from IITP (2022-0-00264/50\%, 2022-0-00951/20\%, 2022-0-00612/20\%, 2020-0-01373/10\%). 

\clearpage

{\small
\bibliographystyle{ieee_fullname}
\bibliography{egbib}
}

\appendix
\section*{Supplementary Material}

This material complements our paper with additional experimental results and their analysis. 
First of all, we verify the solidity of our model, DSR, with the additional quantitative experimental results in Section~\ref{sec: quant}.
Section~\ref{sec: memory} shows the superiority of DSR in terms of memory efficiency.
This is followed by a visual analysis of two modules we proposed, the deformable sampling module and the dependency attention module, in Section~\ref{sec: vis}.
Afterward, Section~\ref{sec: qual} provides some qualitative examples that our model predicts correct answers. 
Lastly, Section~\ref{sec: imple} describes the implementation details, such as the settings for training.
The code will be made publicly available.

\section{Additional Quantitative Results}
\label{sec: quant}

Table~\ref{tab: comp} shows the superior performance of our model, DSR, according to the compositional reasoning step of given questions.
The compositional reasoning step of a question refers to how many steps the question must be inferred in order to find the correct answer. 
In other words, the more reasoning step the question has, the more difficult it is.
As shown in Table~\ref{tab: comp}, our model consistently performs well compared to the strong baseline, ClipBERT, regardless of the compositional steps.
Although our model is designed to target complex questions, it also works effectively for questions with low compositional steps.
In particular, there is a large difference in performance for questions with five compositional steps, which require a lot of spatio-temporal reasoning.

In addition, we validate that the performance gain of DSR comes from novel modules we proposed, but not from the increased parameters, by ablation for each module considering fair parameter sizes. 
Based on the ClipBERT architecture that records the best score in Table 4 in the main paper (a), we add 2 transformer layers in place of the dependency attention, for the text embedding (b).
In addition to (b), we add 4 transformer layers instead of the conditional sampler, for the visual embedding (c).
Remarkably, (b; 52.62) and (c; 52.56) even record lower scores than (a; 53.24).
Simply adding extra parameters does not increase the QA accuracy.
Also, we postulated that adding question embedding to learnable queries is crucial for deformable sampler. Without conditioning on questions, sampled visual tokens will always become identical no matter which questions are given, which would be suboptimal for complex QA tasks. We got an accuracy of 52.39 without the question conditioning, which is lower than our model.
Namely, all the modules we proposed add value.

\begin{table}[t]
\begin{center}
\setlength\tabcolsep{5pt}
\begin{tabular}{lccc}
\toprule
& & DSR(Ours) & ClipBERT~\cite{lei2021less} \\
\midrule
\multicolumn{1}{l}{\multirow{3}{*}{Step 1}} & 
\multicolumn{1}{c}{Binary} & \multicolumn{1}{c}{\textbf{75.24}} & \multicolumn{1}{c}{74.08} \\
& \multicolumn{1}{c}{Open} & \multicolumn{1}{c}{\textbf{9.23}} & \multicolumn{1}{c}{8.46} \\
& \multicolumn{1}{c}{All} & \multicolumn{1}{c}{\textbf{74.98}} & \multicolumn{1}{c}{73.82} \\
\midrule
\multicolumn{1}{l}{\multirow{3}{*}{Step 2}} & 
\multicolumn{1}{c}{Binary} & \multicolumn{1}{c}{\textbf{75.99}} & \multicolumn{1}{c}{75.50} \\
& \multicolumn{1}{c}{Open} & \multicolumn{1}{c}{\textbf{46.92}} & \multicolumn{1}{c}{46.48} \\
& \multicolumn{1}{c}{All} & \multicolumn{1}{c}{\textbf{55.87}} & \multicolumn{1}{c}{55.42} \\
\midrule
\multicolumn{1}{l}{\multirow{3}{*}{Step 3}} & 
\multicolumn{1}{c}{Binary} & \multicolumn{1}{c}{\textbf{79.64}} & \multicolumn{1}{c}{79.61} \\
& \multicolumn{1}{c}{Open} & \multicolumn{1}{c}{\textbf{71.24}} & \multicolumn{1}{c}{70.55} \\
& \multicolumn{1}{c}{All} & \multicolumn{1}{c}{\textbf{74.70}} & \multicolumn{1}{c}{73.87} \\
\midrule
\multicolumn{1}{l}{\multirow{3}{*}{Step 4}} & 
\multicolumn{1}{c}{Binary} & \multicolumn{1}{c}{82.8} & \multicolumn{1}{c}{\textbf{83.82}} \\
& \multicolumn{1}{c}{Open} & \multicolumn{1}{c}{\textbf{49.84}} & \multicolumn{1}{c}{48.86} \\
& \multicolumn{1}{c}{All} & \multicolumn{1}{c}{\textbf{54.01}} & \multicolumn{1}{c}{53.29} \\
\midrule
\multicolumn{1}{l}{\multirow{3}{*}{Step 5}} & 
\multicolumn{1}{c}{Binary} & \multicolumn{1}{c}{\textbf{58.34}} & \multicolumn{1}{c}{48.23} \\
& \multicolumn{1}{c}{Open} & \multicolumn{1}{c}{\textbf{57.78}} & \multicolumn{1}{c}{33.41} \\
& \multicolumn{1}{c}{All} & \multicolumn{1}{c}{\textbf{50.26}} & \multicolumn{1}{c}{38.30} \\
\bottomrule
\end{tabular}
\end{center}
\caption{Quantitative comparison with ClipBERT on the reasoning step based subset of AGQA dataset.}
\label{tab: comp}
%\vspace{-0.5cm}
% \vspace{-0.6cm}
\end{table}

\begin{table*}
    \begin{center}
    \setlength\tabcolsep{4.0pt}
    \begin{tabular}{l|cccccccc|c}
        % \hlineB{2}
        \multicolumn{1}{l|}{} & \multicolumn{8}{c|}{\# of frames} & \multirow{2}{*}{\makecell{Max \\ Frames}} \\ \cline{2-9}
        % \multirow{2}{*}{} & \multicolumn{8}{cl}{\# of frames} & \multirow{2}{*}{\makecell{Max \\ Frames}} \\ \cline{2-9}
        & 2 & 4 & 8 & 16 & 32 & 64 & 128 & 256 & \\ \hline
        Baseline & 7.41 & 7.61 & 8.30 & 10.52 & 16.08 & \textbf{OOM} & \textbf{OOM} & \textbf{OOM} & 60 \\
        ClipBERT~\cite{lei2021less} & 7.39 & 7.91 & 8.86 & 10.14 & 12.71 & 17.84 & 28.15 & \textbf{OOM} & 162 \\
        DSR(Ours) & 6.69 & 6.81 & 7.21 & 8.31 & 9.68 & 13.64 & 23.65 & 32.46 & \textbf{269} \\ 
        % \hlineB{2}
    \end{tabular}
    \end{center}
    \caption{Comparison of memory consumption in GB. Max Frame in the last column means the maximum number of frames right before the OOM error.}
    \label{tab:mem_comp}
    % \vspace{-0.6cm}
\end{table*}

\section{Memory Efficiency of DSR}
\label{sec: memory}
For the spatio-temporally complex QA task, it is important for a model to cover as many frames as possible efficiently. 
Here, we compare how many frames can be addressed by each method until the OOM error is raised under the one NVIDIA V100 GPU environment that has 32GB GPU memory. 
For this experiment, we set the batch size as 1 using the same visual backbone and cross-modal transformer architecture for all methods. For the ClipBERT, each clip consists of 2 consecutive frames, i.e., 64 clips are needed to address 128 frame length, which is the default and the best configuration in their paper~\cite{lei2021less}.

From Table~\ref{tab:mem_comp}, we observe that our DSR shows the best memory efficiency among all comparatives. 
For the Baseline model, all visual features are fed to the cross-modal transformer without any pooling or sparsification. As a result, the memory requirement of the model increases quadratically according to the length of the full visual feature sequence: $\mathcal{O}((THW + L_t)^2)$ where $L_t$ is the length of question words.
In contrast, ClipBERT and DSR can address long sequences more efficiently than the baseline model. In ClipBERT, the feature map is pooled temporally so that only spatial sequence length ($H \times W$) is considered in the cross-modal transformer. 
However, ClipBERT is less efficient than our DSR. Since the ClipBERT passes multiple short clips independently, the memory requirement of the cross-modal transformer becomes $\mathcal{O}(N_c(HW + L_t)^2)$ where $N_c$ denotes the number of clips. In DSR, the memory consumption is only proportional to $N_q + T$: $\mathcal{O}((N_q + T + L_t)^2)$ where $N_q$ and $T$ indicate the number of learnable queries and the length of global context features, respectively.

\section{Visual Analysis on Usefulness of Each Module}
\label{sec: vis}
As introduced in Section 3 of the main paper, we proposed two novel modules for spatio-temporal reasoning. 
To provide the qualitative verification on the effectiveness of each module, this section consists of two parts, 1) qualitative examples of sampled visual features in line with the given question, and 2) visualization of dependency attention weights.

\subsection{Justification of conditionally sampled visual features}

\begin{figure*}
    \begin{center}
    \includegraphics[width=0.95\linewidth]{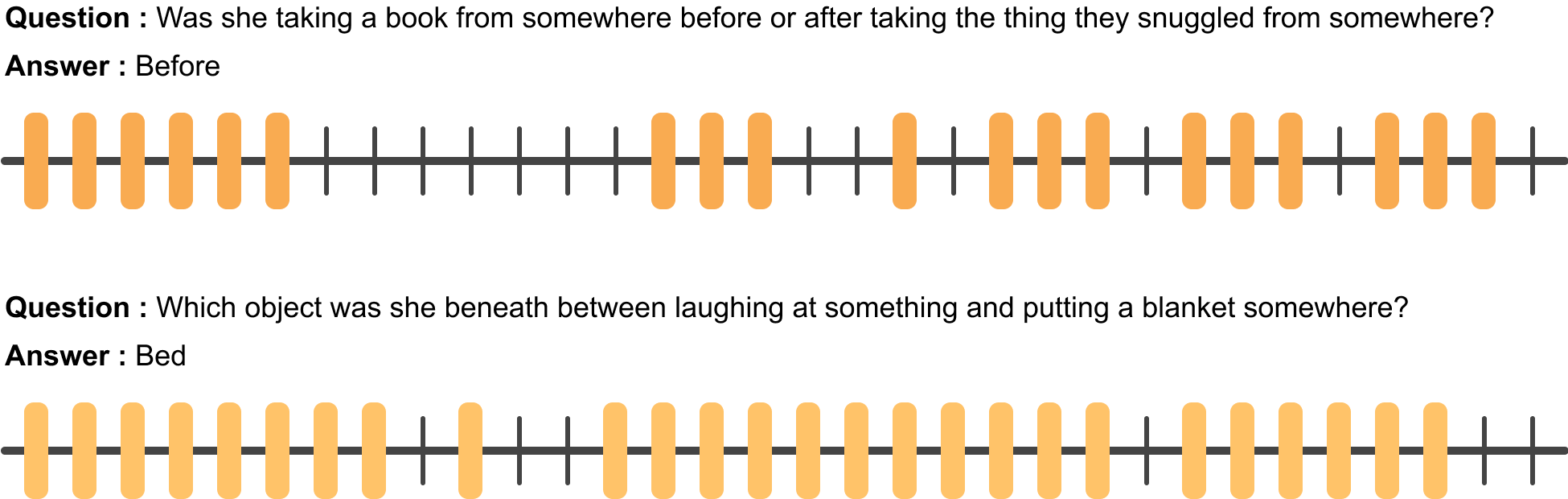}
    \end{center}
    \vspace{-0.4cm}
    \caption{Different sampling points according to different questions in the same video.}
    \label{fig:defa}
    % \vspace{-0.6cm}
\end{figure*}

\begin{figure*}
    \begin{center}
    \includegraphics[width=0.95\linewidth]{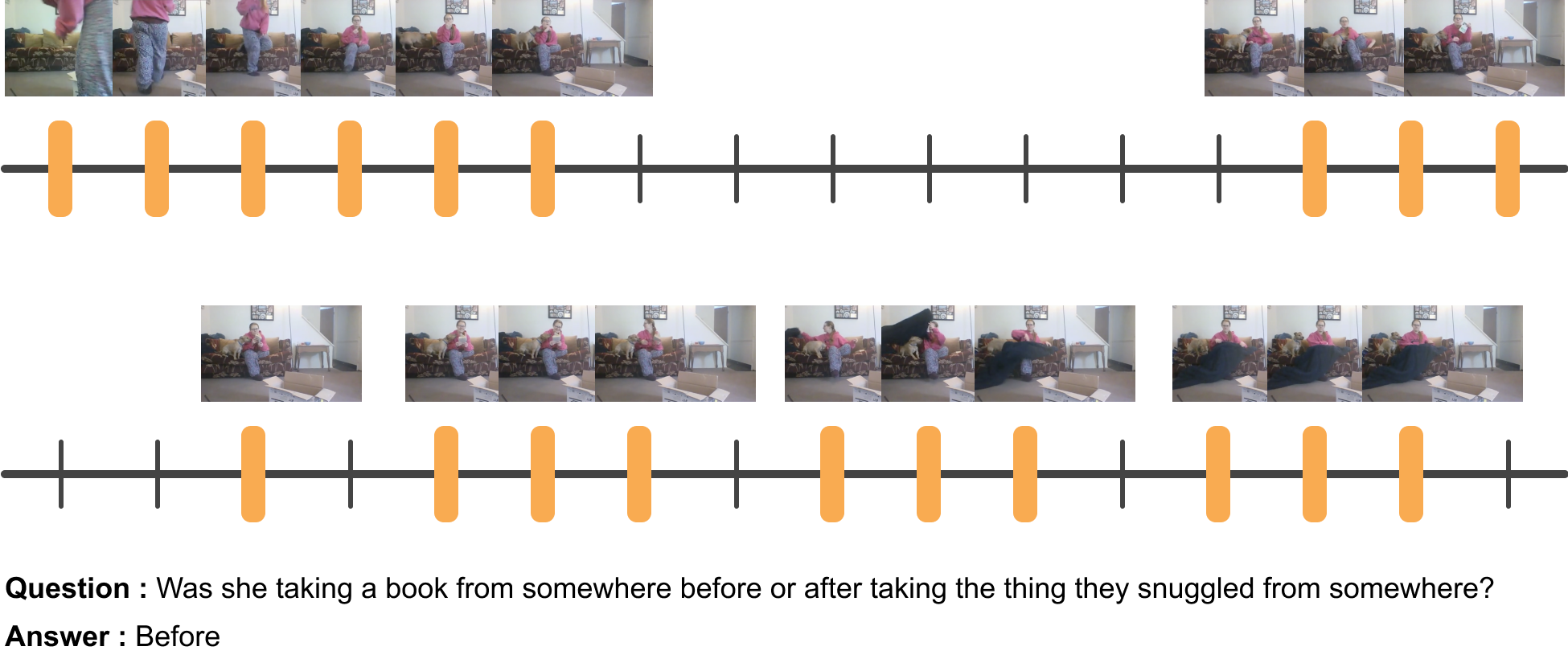}
    \end{center}
    \vspace{-0.6cm}
    \caption{Video frames corresponding to the deformable sampling points.}
    \label{fig:defb}
\end{figure*}

This section justifies the validity of our deformable sampling module, which samples essential visual features conditioned to given questions. 
Instead of densely sampling redundant visual features, the module samples a few diverse samples, which are especially helpful for answering the given questions.
For brevity, Figure~\ref{fig:defa} and~\ref{fig:defb} only represents the sampling points on the temporal axis.
In Figure~\ref{fig:defa}, we observe that our DSR samples different sets of frames based on each question, which indicates that DSR can sample frames in a question conditional way.
Specifically, the model focuses on most of the temporal steps to answer a question in the bottom example of Figure~\ref{fig:defa}, while the top example shows that the model sparsely attends to the specific temporal blocks. 

Figure~\ref{fig:defb} depicts the corresponding video frames along with the sample question. 
To answer the given question, a model should understand the following actions in chronological order; 1) reading a book, 2) taking a blanket, and 3) snuggling under the blanket.
Notably, sampled frames contain all related actions while excluding most of the unnecessary actions.

\subsection{Efficacy of dependency attention head}

\begin{figure*}
\vspace{-0.3cm}
\begin{center}
\includegraphics[width=0.9\linewidth]{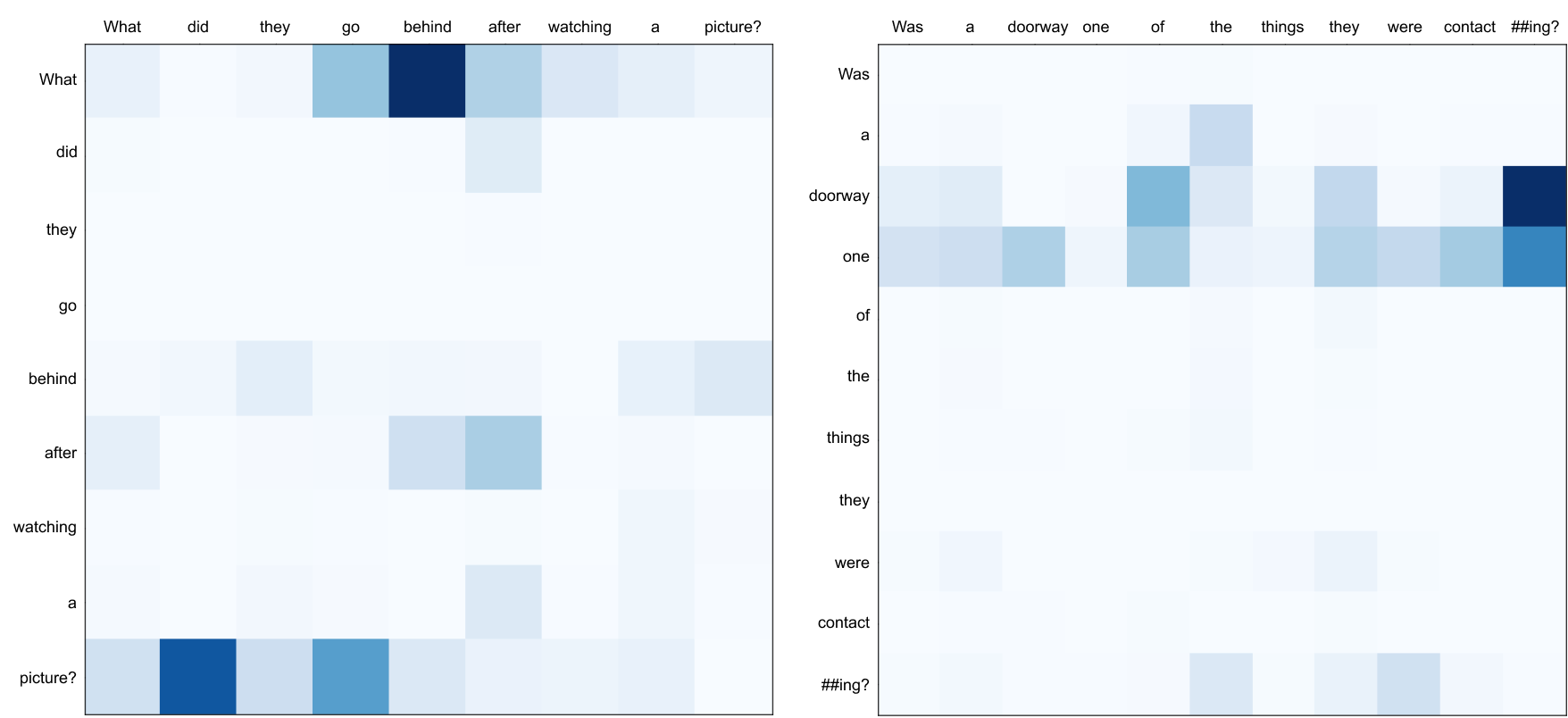}
\end{center}
   \caption{Visualization of self-attention of question tokens that have passed through the dependency attention module.}
\label{fig:dep_example}
\end{figure*}

Figure~\ref{fig:dep_example} visualizes the first head of the last layer of the dependency encoder we proposed, on the test dataset. 
The outputs of the dependency encoder turn into the text inputs of the cross-modal transformer for our model, while the baseline models only utilizes pre-trained Bert embeddings~\cite{devlin-etal-2019-bert} as the input.
Compared to the Bert embeddings that contain general relationships among text tokens, dependency encoder benefits from simplifying long questions by further embedding hierarchical information.
% 예시글 추가하기

\begin{figure*}
    \begin{center}
    \includegraphics[width=0.9\linewidth]{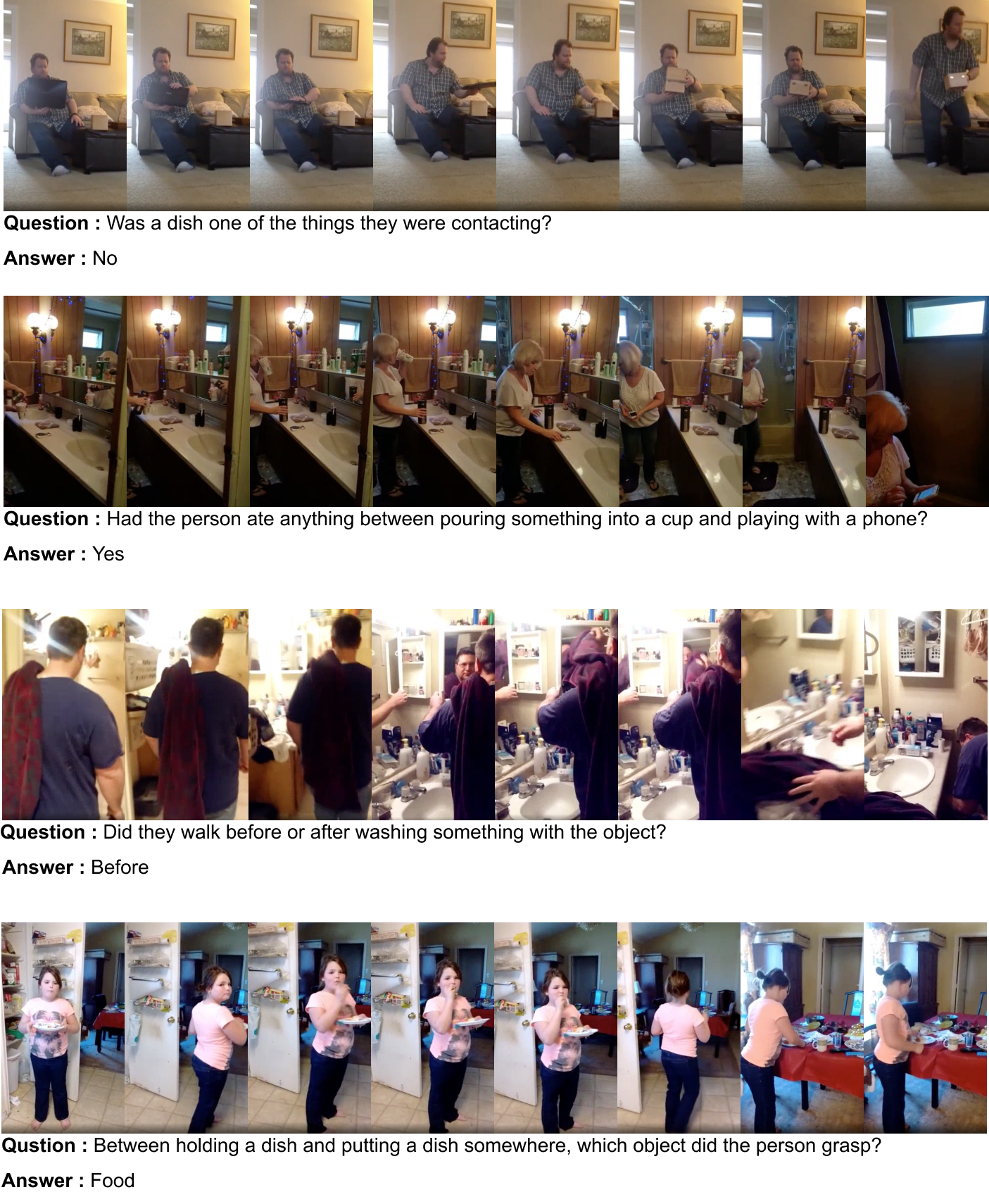}
    \vspace{-0.6cm}
    \end{center}
   \caption{Examples of intricate VideoQA problems of AGQA that our model predicts correct answer.}
\label{fig:qa}
\vspace{-0.6cm}
\end{figure*}

\section{Qualitative Examples of Successful Prediction}
\label{sec: qual}
In Figure~\ref{fig:qa}, we illustrate the examples of AGQA dataset that our model successfully predicts the answer.
As shown in the examples, AGQA dataset consists of problems of solving complex questions for long video sequences containing various actions.
For example, in the case of the third row in Figure~\ref{fig:qa}, the problem can be solved only by recognizing the action of wiping glass and walking while understanding the order of the two actions.
Therefore, our dense but effective model is needed to understand the comprehensive semantic structures of the video.

\section{Implementation Details}
\label{sec: imple}
This section provides the detailed architecture of our method, including the overall framework and two main modules we introduced.
Afterward, we provide the training details, such as hyper-parameters for each objective function, over the dataset we utilized.

\subsection{Model architecture}

\paragraph{Transformer based VideoQA model}
We use the same architecture with ClipBERT~\cite{lei2021less} for the transformer-based VideoQA model. The number of layers and attention heads of each layer is set to 12. The hidden and intermediate dimensions are 768 and 3072, respectively. Also, we use GELU~\cite{hendrycks2016gaussian} action function for the transformer layers. For a classification head, we use 2-fully-connected layers. 

\paragraph{Conditional Deformable Attention module}
For the Conditional Deformable Attention (CDA) module, we use a 4-layer transformer based on the deformable decoder~\cite{zhu2020deformable}. In each layer, we sample 8 offset points with 4 different attention heads. The hidden dimension of each transformer layer is 768 and we use ReLU~\cite{nair2010rectified} as an activation function.

\paragraph{Dependency attention module}
The 2-layer transformer with 12-multi-head is used as a backbone of the dependency attention module. The dependency-constrained attention module is implemented on the first head of the first layer. For the dependency head, attention probabilities are calculated based on dependency parsing relations and learnable value embeddings are multiplied by corresponding attention probabilities.

\subsection{Hyperparameters and training details}
\paragraph{Default Hyperparameters and Optimization}
For all experiments by default, we set the learning rate and weight decay for all modules except for the CDA to 5e-5 and 1e-3, respectively. Also, we use AdamW~\cite{loshchilov2017decoupled} optimizer and use learning rate warmup over the first 10\% training steps followed by linear decay to 0. 
Following the optimization strategy for \cite{zhu2020deformable}, the base learning rate of CDA module is 0.0001 and the learning rate is decayed at half of the training steps by a factor of 0.1. Also, learning rates of the linear projections, used for predicting object query reference points and sampling offsets, are multiplied by a factor of 0.1.
For the video inputs, we resize frames to 448 pixels for the long spatial side and add zero padding to the remaining regions for the short side. 
Also, we set the maximum question length to 100 for all experiments except for TGIF-QA. For the TGIF-QA the maximum question length is 25.

\paragraph{Training details}
For our main AGQA experimental results, we train our DSR for 5 epochs with a learning rate of 2e-4. We use a total of 32 NVIDIA V100 GPUs with a batch size of 8 per GPU. For ablations, we use 4 GPUs with a learning rate of 5e-5. 
For MSRVTT-QA and TVQA, we train our model for ten epochs and five epochs, respectively, and the remaining training details are the same as the main AGQA experiment.
For TGIF-QA, we train ClipBERT and DSR for 60 epochs with a dropout probability of 0.4 for the final classification head. Other hyperparameters such as learning rate and weight decay are the same as the default setting described above.

\end{document}